\documentclass{article} 
\usepackage{iclr2023_conference,times}

\usepackage{booktabs}
\usepackage{graphicx}
\usepackage{wrapfig}
\usepackage{algorithm,algpseudocode}
\usepackage{multicol}
\usepackage{multirow}
\usepackage{authblk}

\algdef{SE}[DOWHILE]{Do}{doWhile}{\algorithmicdo}[1]{\algorithmicwhile\ #1}%
\algnewcommand\algorithmicforeach{\textbf{for each}}
\algdef{S}[FOR]{ForEach}[1]{\algorithmicforeach\ #1\ \algorithmicdo}

\usepackage{amsmath,amsfonts,bm}

\def\eqref#1{equation~\ref{#1}}

\def\1{\bm{1}}

\DeclareMathAlphabet{\mathsfit}{\encodingdefault}{\sfdefault}{m}{sl}
\SetMathAlphabet{\mathsfit}{bold}{\encodingdefault}{\sfdefault}{bx}{n}

\usepackage[hidelinks]{hyperref}
\usepackage{url}
\usepackage{subfig}

\newcommand{\round}[1]{\ensuremath{\lfloor#1\rceil}}

\title{FPTQ: Fine-grained Post-Training Quantization for Large Language Models}

\iclrfinalcopy

\author[1]{Qingyuan Li $^\dagger$}
\author[1,2]{Yifan Zhang $^\dagger$\thanks{Work done as an intern at Meituan. $^\dagger$ Equal Contribution.}}
\author[1]{Liang Li}
\author[1]{Peng Yao}
\author[1]{Bo Zhang}
\author[1]{Xiangxiang Chu}
\author[1]{Yerui Sun}
\author[2]{Li Du}
\author[1]{Yuchen Xie}
\affil[1]{Meituan}
\affil[2]{Nanjing University}

\begin{document}

\maketitle

\begin{abstract}
In the era of large-scale language models, the substantial parameter size poses significant challenges for deployment. Being a prevalent compression technique, quantization has emerged as the mainstream practice to tackle this issue, which is mainly centered on two recipes W8A8 and W4A16 (i.e. weights and activations in such bit widths). 
In this study, we propose a novel W4A8 post-training quantization method for the available open-sourced LLMs, which combines the advantages of both two recipes. Therefore, we can leverage the benefit in the I/O utilization of 4-bit weight quantization and the acceleration due to 8-bit matrix computation. Nevertheless, the W4A8 faces notorious performance degradation. As a remedy, we involve layerwise activation quantization strategies which feature a novel logarithmic equalization for most intractable layers, and we combine them with fine-grained weight quantization. Without whistles and bells, we eliminate the necessity for further fine-tuning and obtain the state-of-the-art W4A8 quantized performance on BLOOM, LLaMA, and LLaMA-2 on standard benchmarks. We confirm that the W4A8 quantization is achievable for the deployment of large language models, fostering their wide-spreading real-world applications.
\end{abstract}

\section{Introduction}

Large Language Models (LLMs) are distinguished for their exceptional \emph{emergent knowledge capacity} \citep{wei2022emergent}, enabling them to perform admirably across a wide variety of language tasks. However, their massive scale poses a significant hurdle to deployment due to the substantial storage and the huge amount of computation required. This challenge is particularly pronounced in environments with limited resources such as edge computing devices and personal devices, where the constraints can inhibit the widespread adoption  of these cutting-edge language models.

To address this issue, several model compression strategies have been proposed, including pruning~\citep{ma2023llmpruner,frantar2023sparsegpt,sun2023simple}, distillation~\citep{zhang2023lifting}, quantization~\citep{frantar2022gptq,xiao2023smoothquant}, and low-rank decomposition \citep{yao2023zeroquantv2}. Each of these approaches has its own limitations. For instance, pruning can achieve reasonable compression rates but it may require significant fine-tuning or are closely tied to specific hardware architectures. In contrast, quantization techniques, despite their universal applicability, are often confronted with the problem of significant quantization errors, particularly with the increasing parameter sizes \citep{dettmers2022llm}.

Lately, research attention has been shifted towards a more balanced approach to quantization, specifically the usage of lower-bit widths for weights and higher-bit widths for activation, like W4A16 in GPTQ~\citep{frantar2022gptq}. This introduces a novel perspective to tackle the computational and memory-intensive aspects of LLMs, which are typically composed of Transformer Decoder structures ~\citep{vaswani2017attention}. During inference, it can be divided into compute-intensive \emph{context decoding} stage and memory-intensive \emph{self-decoding} stage, each presenting unique challenges and opportunities for further optimization.

However, there is a conspicuous dearth of research that explores the synergistic combination of two quantization recipes W8A8 and W4A16. This paper aims to bridge the gap by proposing an innovative Fine-grained Post-Training Quantization (called FPTQ) method that combines the benefits of both, thereby providing an effective and efficient W4A8 solution for the deployment of a variety of available large language models that are tested across a myriad of natural language tasks.


\begin{figure*}[htbp]
  \centering
    \includegraphics[width=\textwidth, trim=200 0 200 0]{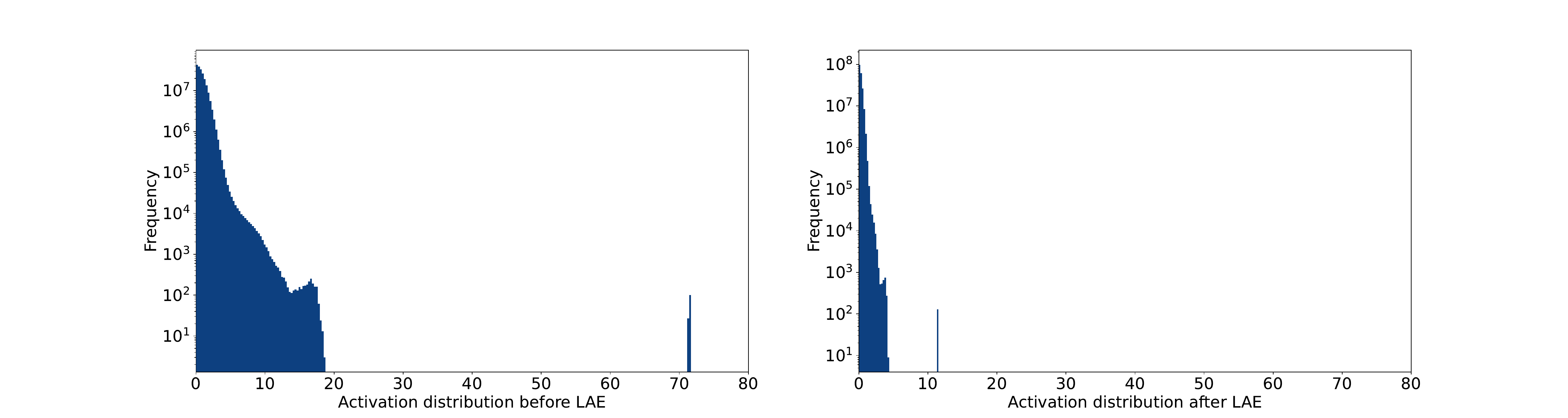}
  \caption{Activation distribution before and after logarithmic equalization on BLOOM-7B1.}
  \label{fig:contrast of Activation distribution after equlizatin}
\end{figure*}

We first investigate the quantization difficulty by illustrating the activation distributions in different layers, discovering that their ranges differ dramatically which motivates us for a layerwise strategy. Subsequently, we provide a unique activation equalization technique to handle the intractable outliers (Figure~\ref{fig:contrast of Activation distribution after equlizatin}), and improve the overall performance with fine-grained weight quantization. 

In a nutshell, we make several key contributions to the field of LLM compression and deployment:

\begin{enumerate}  
    \item \textbf{High performance and low-cost W4A8 compression}: We are the \emph{first} to achieve high-performance W4A8 (INT4 weights and INT8 activation) PTQ compression for large language models, maintaining the accuracy of the original model. Being a post-training quantization technique, it tremendously simplifies the production flow of LLMs.
    \item \textbf{Novel quantization scheme}: Based on our comprehensive analysis of the activation distribution of LLMs, we employ a layerwise strategy to cope with different levels of quantization difficulty. Particularly, we devise an offline \emph{logarithmic activation equalization} to render a quantization-friendly distribution for previously intractable layers. 
    \item \textbf{Inference-friendly}: Our approach harmonizes the memory and computation efficiency which enables the storage of weights in a 4-bit format while executing INT8 inference, thereby catalyzing both the memory access and computation.
\end{enumerate}









\section{Related Work}





\subsection{Large Language Models}
The past few years have witnessed the booming of pre-trained language models. BERT~\citep{devlin2018bert} is designed to understand the context of words in a sentence and has been used for tasks such as sentiment analysis and question answering. RoBERTa~\citep{liu2019roberta} is an improved version of BERT with better pre-training techniques and larger training data. T5~\citep{raffel2020exploring} is designed to perform a wide range of natural language processing tasks, including language translation and summarization. XLNet~\citep{yang2019xlnet} is designed to handle long sequences of text and has achieved state-of-the-art results on several natural language processing tasks. GPT-3~\citep{brown2020language} is one of the most advanced LLMs with 175 billion parameters, capable of performing a wide range of natural language processing tasks. Along with the open-sourced ones like GLM \citep{du2021glm}, BLOOM \citep{laurencconbigscience}, OPT \citep{zhang2022opt} and LLaMa series  \citep{touvron2023llama}, LLMs have remarkably revolutionized the field of natural language processing and are being used in a wide range of applications.

Nevertheless, LLMs have billions of parameters and are often pre-trained on large amounts of text data, which require significant computational resources to train and deploy. There is a call for faster inference time and lower memory requirements to make LLMs more practical.

\subsection{Quantization on LLMs}

Applying quantization to large language models presents unique challenges. Traditional PTQ schemes have achieved great success in Convolutional Neural Networks (CNN)~\citep{nagel2019data,wu2020integer,nagel2021white,yao2021hawq}, but direct application to large language models often results in severe accuracy loss, this is typically due to the presence of many outliers in the activation values of large models~\citep{dettmers2022llm}. 

Several approaches have been proposed to address these issues. For example, LLM.int8()~\citep{dettmers2022llm} splits the input activation values into two parts: non-outlier dimensions computed with INT8, and outliers computed with FP16. GPTQ~\citep{frantar2022gptq} and AWQ~\citep{lin2023awq} circumvent this difficulty by adopting FP16 activation and INT4 weight-only quantization. However, these methods also have their limitations, such as computational overhead and the inability to truly leverage hardware acceleration.

Other approaches like SmoothQuant ~\citep{xiao2023smoothquant}, RPTQ ~\citep{yuan2023rptq}, and ZeroQuant-V2 ~\citep{yao2023zeroquantv2} propose different strategies to achieve quantization while mitigating the accuracy loss and computational overhead. However, SmoothQuant is merely a W8A8 solution and it suffers from poor performance on W4A8. The rest tackles the W4A8 challenge but they come with their own set of challenges like weight reordering, asymmetric quantization, and group-wise activation, which can perplex the engineering work and may not well facilitate hardware. 
In light of these problems, we are driven to achieve W4A8 quantization without relying on QAT or distillation methods, paving the way for the efficient deployment of LLMs.

\section{Method}




\subsection{Why W4A8?}
\begin{figure}[ht]
\centering
  \includegraphics[width=\textwidth]{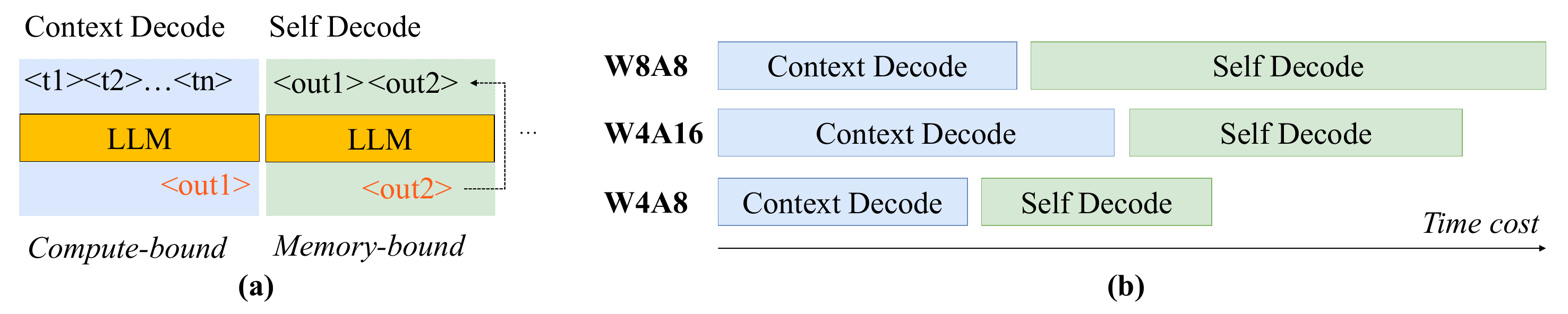}
  \caption{\textbf{(a)} Two stages of LLM inference where context decoding is compute-bound and self-decoding is memory-bound. \textbf{(b)} W4A8 speeds up both stages and is faster than the other two.}
  \label{fig:w4a8-motivation}
\end{figure}

The generative inference of LLMs can be divided into two stages: \emph{context decoding} that generates an output token given an input prompt (embedded as a sequence of tokens), and \emph{self-decoding} that iteratively predicts the next token in a sequence, see Figure~\ref{fig:w4a8-motivation} (a). The former is compute-bound due to the first-round computation of lengthy input sequences and the latter is memory-bound as a result of sequential processing, thus two different implementations are required. 

\begin{wraptable}{r}{8cm}
  \vskip -0.2in
  \small
  \centering
  \caption{Comparison of decoding stage efficiency for different quantization methods. CD: Context Decoding, SD: Self-Decoding}
  \vskip -0.1in
    \begin{tabular}{lcc}
    \toprule
    Method & Efficient CD  & Efficient SD \\
    \midrule
    ZeroQuant & No & No \\
    SmoothQuant & Yes & No \\
    GPTQ & No & Yes \\
    AWQ & No & Yes \\
    Ours & \textbf{Yes} & \textbf{Yes} \\
    \bottomrule
    \end{tabular}
  \label{tab:decoding-efficiency}
  \vskip -0.3in
\end{wraptable}

Previous quantization methods like Smoothquant~\citep{xiao2023smoothquant} features W8A8, while AWQ~\citep{lin2023awq} and GPTQ~\citep{frantar2022gptq} use W4A16. Both recipes compromise one stage for another, leading to inferior overall performance, whereas only W4A8 can boost both stages, see Figure~\ref{fig:w4a8-motivation} (b) and Table~\ref{tab:decoding-efficiency}. That is, context decoding enjoys the speed-up using 8-bit matrix multiplication, while self-decoding is also accelerated via reduced memory access using 4-bit weight.

There are a few existing W4A8 studies. ZeroQuant~\citep{yao2022zeroquant} utilizes mixed precision for self-attention weights (W8) and is not tested on larger models, ZeroQuantV2~\citep{yao2023zeroquantv2} uses fine-grained activation quantization which is not feasible in practice. ZeroQuant-FP \citep{wu2023zeroquantfp} alleviates the degradation by using higher-precision FP8 computation but it depends on specific hardware (e.g. NVIDIA H100). LLM-QAT~\citep{liu2023llm} adopts QAT to improve W4A8 performance but it requires costly training and is prone to tedious hyper-parameter tuning. Therefore, it is necessary to improve the accuracy of the W4A8 model while not harming its inference speed. The method shall also be made low-cost and generalizable for most up-to-date LLMs.




\subsection{Analysis of activation distribution on LLMs}

\begin{figure*}[htbp]
  \centering
    \includegraphics[width=\textwidth, trim=200 0 200 0]{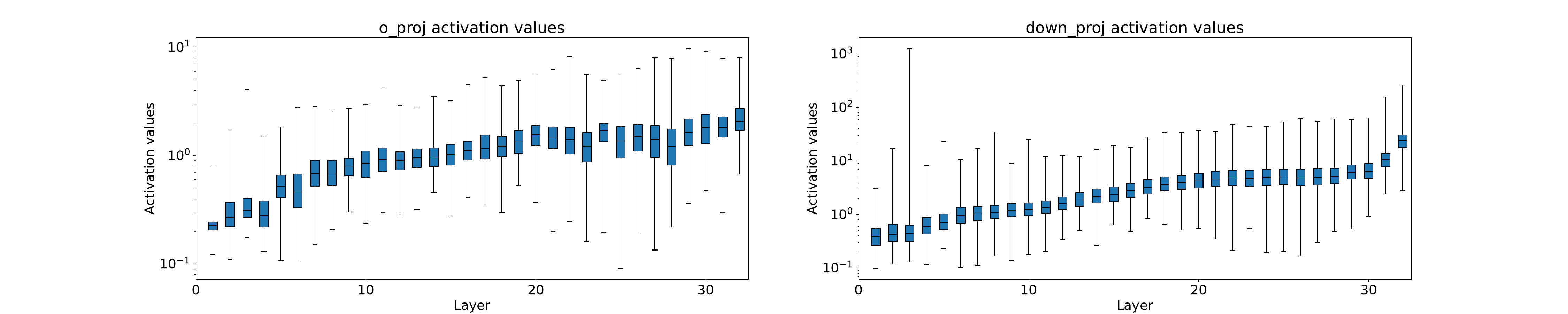}
  \caption{Visualization of activation distribution of $o_{proj}$ and $down_{proj}$ on LLaMA-7B.}
  \label{fig:activation distribution on LLMs}
\end{figure*}

With our goal in mind, we are driven to design a robust PTQ method. To begin with, we study why vanilla W4A8 quantization is difficult for current LLMs. We first draw the activation distribution of LLaMA-7B in Figure~\ref{fig:activation distribution on LLMs} to find the distinct behaviors of different layers. For instance, $o_{proj}$ has compact distribution while $down_{proj}$ spans extensively. This phenomenon reoccurs in many other LLMs, see Appendix~\ref{app:more-act-dist}.

As we can see from the above analysis, the maximum fluctuation range of input activation values for certain layers ranges from tens to thousands. Using \emph{per-tensor static quantization} will result in significant quantization errors, but using \emph{per-token dynamic quantization} for all layers will not bring adequate hardware acceleration. Therefore, it naturally calls for a layer-specific policy to determine the granularity of quantization.

\subsection{FPTQ: Fine-grained Post-training Quantization}

Motivated by the above analysis, we propose our post-training quantization method which employs a layerwise quantization strategy regarding disparate activation distributions. Our complete procedure is given in Algorithm~\ref{alg:FPTQ}. The key components are discussed in detail.

\begin{algorithm}[ht]
\caption{FPTQ: Fine-grained Post-Training Quantization} 
\label{alg:FPTQ}
\textbf{Input:} A pre-trained LLM \\
\textbf{Output:} A quantized LLM
\begin{algorithmic}[1]
\State Calibrate the pre-trained LLM with a predefined dataset
\State Perform activation distribution analysis
\ForEach{layer-$l$ in  the Transformer structure ($L$ layers in total)}
\If {Activation range $ v \leq v_0 $ }
    \State Set layer $l$'s activation quantization policy to \emph{static per-tensor}
\ElsIf {Activation range $ v_0 < v < v_1 $} 
    \State Perform \emph{logarithmic activation equalization}
    \State Set layer $l$'s activation quantization policy to \emph{static per-tensor}
\Else  
    \State Set layer $l$'s activation quantization policy to \emph{dynamic per-token}
\EndIf 
\State Set each layer's weight quantization policy as fine-grained 
\EndFor
\State Update the LLM's weights and activations w.r.t. the chosen quantization policy
\State Get the high-performance quantized LLM
\end{algorithmic}
\end{algorithm}

\subsubsection{Layer-wise Activation Quantization Strategy}
The key to resolving the activation quantization difficulty lies in the outlier treatment. 
Empirically, we can use different activation quantization strategies for different layers, as shown in  Table~\ref{tab:quantization-strategies}. For activation value ranges within tens (denoted as $v_0$), per-tensor static quantization can be safely used. However, to avoid quantization loss for activation ranges over hundreds (denoted as $v_1$), per-token dynamic quantization shall be put in place although slightly sacrificing hardware acceleration benefits. For most layers that range within hundreds, i.e. $(v_0, v_1)$, it demands a particular strategy that simultaneously reduces the quantization error while not harming the inference speed.

\begin{table}[htbp]
\setlength\tabcolsep{4pt}
  \centering
  \caption{Activation quantization strategies for different ranges of activation values.} 
    \begin{tabular}{cccc}
    \toprule
    \textbf{Activation Value Range} & \textbf{Quantization Strategy} & \textbf{Hardware Efficiency} & \textbf{Typical Operation} \\
    \midrule
    $v \leq v_0 $  & per-tensor, static  & High & Dense \\
    $v_0 < v < v_1$ & LAE + per-tensor, static  & High & QKV, FC1\\
    $v \geq v1$ &  per-token, dynamic & Medium  & FC2 \\
    \bottomrule
    \end{tabular}%
  \label{tab:quantization-strategies}%
\end{table}%

\cite{xiao2023smoothquant} discover that when larger outliers dominate the distribution, the effective quantization bits of inliers are substantially narrowed. For per-tensor 8-bit quantization, it becomes  $2^8 \cdot m_i/ m$  where $m_i$ is the maximum amplitude of channel $i$ and $m$ is the maximum value of the whole tensor. They also observe that outliers stay in fixed channels. Based on these two findings, we are allowed to perform per-channel outlier suppression on activations. SmoothQuant~\citep{xiao2023smoothquant} attempts to `smooth' per-channel distribution by dividing the activation with a scale $\mathbf{s}_i = \max(|\mathbf{x}_i|) / \max(|\mathbf{w}_i|)$, where $\mathbf{x}_i$ and $\mathbf{w}_i$ are activation and weight of channel $i$ respectively.  AWQ~\citep{lin2023awq} introduces grid-searched hyper-parameters $\alpha$ and $\beta$ to lay importance to activation and weight separately, where they find the contribution of weights is marginal and suggest activation-awareness is most important.
In this paper, we argue that it is unnecessary to consider weights for computing the activation ``smoothing" scale. Besides, it is crucial to retain all the activation values with a \emph{non-linear lossless mapping}, yet it has to satisfy two criteria \textbf{(1)} touching gently with the inliers \textbf{(2)} harshly suppressing the outliers. In this regard, we verify that the logarithmic function rightly fits this purpose. 

\textbf{Logarithmic Activation Equalization.} To render a quantization-friendly activation distribution, we propose a new offline \emph{logarithmic activation equalization} (LAE) method that moderates activation distributions in a non-linear fashion. Specifically, we compute the $i$-th channel scale $\mathbf{s}_i$ as the maximum activation value $\max(|\mathbf{X}_i|)$ divided by its logarithmic mapping with a shift of 2 (to have a minimum of scale 1), shown in Equation~\ref{eq:log-act-eq}. The formula retains the original information while it squashes various distributions comparably. Figure~\ref{fig:contrast of Activation distribution after equlizatin} exhibits its outcome distribution. 

\begin{align}\label{eq:log-act-eq}
    \mathbf{s}_i &= \max(|\mathbf{x}_i|) / \log_2 (2 + \max(|\mathbf{x}_i|)); \quad \mathbf{x}_i = \mathbf{x}_i / \mathbf{s}_i
\end{align}

Once the scale $\mathbf{s}$ is obtained, we can update the corresponding weight and activation as follows,

\begin{align}\label{eq:weight-act-update}
\mathbf{W'} = \text{diag}(\mathbf{s}) \mathbf{W}; \quad \mathbf{X'} = \mathbf{X} \text{diag}(\mathbf{s})^{-1} \quad s.t. \quad \mathbf{X'} \mathbf{W'}  =  \mathbf{X} \mathbf{W} 
\end{align}

Hence, this update is made \emph{in-place} as it is mathematically equivalent. Notably, $\mathbf{s}$ can be easily fused into the weight of the previous layer. In our case, there are only two types of operations (QKV and FC1) whose activation ranges are in ($v_0, v_1$). To apply the offline LAE, their activation updates are fused into their preceding operation LayerNorm~\citep{ba2016layer}.

\subsubsection{Weight Quantization} 

\begin{figure}[ht]
\centering
  \includegraphics[width=\textwidth]{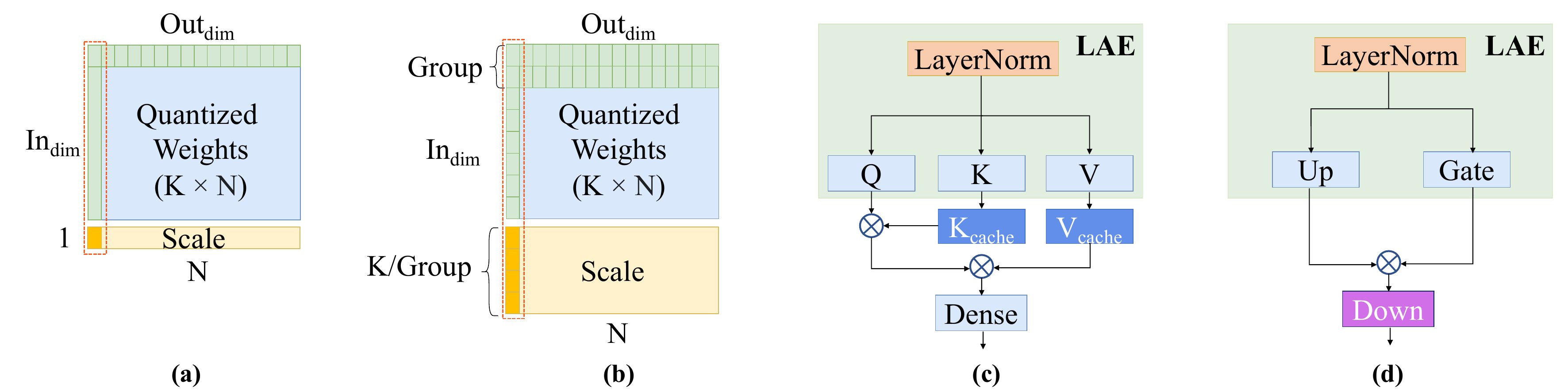}
  \caption{\textbf{(a)} Per-channel weight quantization. \textbf{(b)} Fine-grained per-channel quantization. \textbf{(c, d)} Self-attention and FFN in most LLMs. Light blue: per-tensor static activation quantization. Purple: per-token dynamic activation quantization. All weights are quantized in a fine-grained manner.}
  \label{fig:per-channel-finegrained}
\end{figure}

Due to the intricacies of LLMs, it is not tractable to use the vanilla per-channel strategy only, as shown in Figure~\ref{fig:per-channel-finegrained} (a). ZeroQuant~\citep{yao2022zeroquant} adopts a fine-grained groupwise weight quantization~\citep{shen2020q} that addresses the quantization difficulty of smaller LLMs like GPT-3 \citep{brown2020language}. As the two strategies are identically costly from the engineering perspective, we adopt fine-grained weight quantization where the scale is computed groupwise for all layers to obtain better performance, depicted in Figure~\ref{fig:per-channel-finegrained} (b). 

As LLaMA series~\citep{touvron2023llama} rises to the mainstream focus, we illustrate our specific quantization scheme for its architecture in Figure~\ref{fig:per-channel-finegrained} (c) and (d). Interestingly, we discover that the trend of LLaMA activation distributions holds for all model series, such that our quantization scheme can be directly reused. Logarithmic activation equalization is performed offline (the scale for activation is then fused into LayerNorm) for QKV and Up/Gate. It's also worth noting that the quantized KV cache is applied to save I/O costs.

\section{Experiment}

\subsection{Datasets}

We validated our quantization scheme on several datasets, including LAMBADA \citep{paperno2016lambada}, MMLU~\citep{hendrycks2020measuring}, and a set of Common Sense QA~\citep{talmor2019commonsenseqa} tasks like WinoGrande~\citep{sakaguchi2021winogrande}, PIQA~\citep{tata2003piqa}, HellaSwag~\citep{zellers2019hellaswag}, ARC$_e$. For CommonSense QA tasks, we used the Language Model Evaluation Harness~\citep{eval-harness} tool to evaluate our models. For the calibration set, we randomly sampled 512 samples from the Pile dataset~\citep{gao2020pile}.

\begin{wraptable}{r}{8cm}
\vskip -0.2in
\setlength\tabcolsep{2pt}
\centering
\begin{tabular}{l c c c}
\toprule
\textbf{Model} & \textbf{Original} & \textbf{SmoothQuant} & \textbf{FPTQ} \\ 
 & FP16 & W8A8 & W4A8 \\
\midrule
BLOOM-7B1 & 57.9080\% & 59.6352\% & 58.2185\% \\ 
\midrule
LLaMA-7B & 73.7435\% & 73.7823\% & 73.8017\% \\ 
LLaMA-13B & 76.1886\% & 76.3633\% & 75.7423\% \\
LLaMA-65B & 79.1966\% & 78.6920\% & 79.1384\% \\ 
\midrule
LLaMA-2-7B & 73.7046\% & 74.1510\% & 72.4820\% \\ 
LLaMA-2-13B & 76.6350\% & 75.5288\% & 75.3154\% \\ 
LLaMA-2-70B & 79.5653\% & 78.7891\% & 78.7114\% \\ 
\bottomrule
\end{tabular}
\caption{Comparison on the LAMBADA Dataset.}
\label{tab: lambada acc}
\end{wraptable}

\subsection{Implementation}
\textbf{Baselines.} In our experiments, we selected SmoothQuant \citep{xiao2023smoothquant} and GPTQ~\citep{frantar2022gptq} as our baselines, given their status as the most prevalent W8A8 and W4A16 quantization schemes, respectively. These methods have been widely adopted in various applications and their performance has been extensively validated, establishing them as reliable benchmarks in the field of LLMs quantization.
Simultaneously, to further demonstrate the potential of FPTQ, we compare it with the QAT method, particularly with LLM-QAT~\citep{liu2023llmqat}. It's worth mentioning that QAT introduces a significant computational resource overhead; in contrast, our approach incurs a negligible cost compared to it.

\textbf{Implementation.} We find that for the investigated LLMs in our paper, the activation bound $v_0$ can be typically set as 15 and $v_1$ 150.


\subsection{Experimental Results on LAMBADA}

We initially conducted our experiments on the LAMBADA dataset~\citep{paperno2016lambada}. Despite the fact that LAMBADA may not effectively reflect the comprehensive capabilities of the model, it serves as a valuable tool for rapidly validating model precision and quantifying the impact on model performance. Our method, Fine-grained Post-training Quantization (FPTQ), achieved W4A8 quantized models that demonstrated precision strikingly similar to their floating-point counterparts on both the BLOOM-7B1~\citep{scao2022bloom} and all models in the LLaMA series~\citep{touvron2023llama}. This is a highly encouraging observation, suggesting the efficacy of our approach.

\begin{table}[ht!]
\centering
\small
\setlength\tabcolsep{1pt}
\begin{tabular}{@{}l*{12}{c}}
\toprule
Model & \multicolumn{2}{c|}{HyperParam}  & \multicolumn{5}{c|}{MMLU} & \multicolumn{5}{c}{Common Sense QA} \\
& Method & BW & Hums. & STEM & Social & Other & Avg & WG & PIQA & HS & ARC$_e$ & Avg \\
\midrule
BLOOM-7B1 & FP16 & W16A16 & 26.10 & 26.84 & 24.21 & 26.34 & 25.90 & 63.93 & 72.91 &	57.24 & 57.74 &	62.96 \\ 
 & SmoothQuant & W8A8 & 26.04 & 27.80 & 24.50 & 25.82 & 26.03 & 61.96 & 72.52 & 56.66 &	57.41 &	62.14\\ 
 & GPTQ & W4A16 & 26.06 & 26.47 & 25.28 & 26.50 & 26.08 & 63.38 & 72.42 & 55.98 & 56.86 & 62.16 \\
 &FPTQ & W4A8 & 25.87 & \textbf{26.71} & 23.76 & \textbf{26.56} & 25.74 & \textbf{63.22} & \textbf{72.80} & 55.98 & \textbf{57.32} & \textbf{62.33} \\
\midrule
LLaMA-7B & FP16 & W16A16 & 33.60 & 31.10 & 38.20 & 38.40 & 35.20 & 69.85 & 79.16 & 76.10 & 72.80 & 74.48 \\
  &SmoothQuant& W8A8 & 33.88 & 30.32 & 37.63 & 39.08 & 35.14 & 70.09 & 79.00 & 75.17 & 72.22 & 74.12 \\
  & GPTQ & W4A16 & 32.39 & 30.35 & 35.03 & 36.15 & 33.40 & 68.03 & 77.69 & 72.95 & 69.44 & 72.02 \\
 & FPTQ & W4A8 & 30.20 & 29.95 & 32.76 & 35.87 & 32.02 & 70.01 & 78.40 & 74.46 & 70.79 & 73.42 \\
\midrule
LLaMA-13B & FP16 & W16A16 & 44.60 & 37.10 & 54.00 & 53.50 & 47.10 & 72.77 & 80.09 & 79.07 & 74.71 & 76.66 \\
&SmoothQuant& W8A8 & 44.14 & 36.51 & 54.05 & 52.65 & 46.64 & 72.06 & 79.71 & 78.34 & 73.91 & 76.00 \\
 & GPTQ & W4A16 & 46.01 & 39.00 & 54.01 & 53.36 & 47.96 & 73.16 & 80.25 & 78.60 & 74.37 & 76.59 \\
 & FPTQ & W4A8 & 40.96 & 34.19 & 49.72 & 49.75 & 43.46 & 72.14 & 79.33 & 77.50 & 72.69 & 75.41 \\
\midrule
LLaMA-65B & FP16 & W16A16 & 61.80 & 52.00 & 73.30 & 67.60 & 63.50 & 77.35 & 82.32 & 84.15 & 79.76 & 80.90 \\
 &SmoothQuant& W8A8 & 61.32 & 50.50 & 71.69 & 66.90 & 62.56 & 74.90 & 81.07 & 82.32 & 77.4 & 78.92 \\
 & GPTQ & W4A16 & 60.23 & 52.09 & 72.15 & 66.75 & 62.60 & 77.43	& 82.32 & 83.57 & 79.88 & 80.80 \\
 & FPTQ & W4A8 & 59.85 & 49.24 & 71.50 & 65.89 & 61.52 & 75.77 & \textbf{81.45} & \textbf{83.44} & \textbf{78.45} & \textbf{79.78} \\
\midrule
 LLaMA-2-7B & FP16 & W16A16 & 43.40 & 37.00 & 51.80 & 52.40 & 46.00 & 69.06 & 79.11 & 75.98 & 74.58 & 74.68 \\ 
 &SmoothQuant& W8A8 & 42.49 & 36.65 & 50.67 & 51.33 & 45.06 & 69.06 & 77.97 & 75.91 & 75.98 & 74.58 \\ 
 & GPTQ & W4A16 & 42.66 & 36.45 & 51.25 & 50.99 & 45.13 & 68.51 & 78.67 & 0.75.96 & 71.68 & 73.45 \\ 
 & FPTQ & W4A8 & 41.15 & 35.79 & 49.37 & 50.77 & 44.02 & \textbf{69.38} & 77.97 & 74.89 & 72.85 & \textbf{73.77} \\ 
\midrule
LLaMA-2-13B & FP16 & W16A16 & 54.40 & 44.30 & 63.40 & 60.80 & 55.70 & 72.22 & 80.52 & 79.38 & 77.44 & 77.39 \\ 
 &SmoothQuant& W8A8 & 52.67 & 43.07 & 63.15 & 60.39 & 54.69 & 72.06 & 79.54 & 79.28 & 77.31 & 77.05 \\ 
 & GPTQ & W4A16 & 51.99 & 43.57 & 63.05 & 60.49 & 54.56 & 72.30 & 79.60 & 78.79 & 77.23 & 76.98 \\ 
 & FPTQ & W4A8 & 51.65 & 42.54 & 62.27 & 59.90 & 53.92 & 70.56 & 79.43 & 78.06 & 75.76 & 75.95 \\ 
\midrule
LLaMA-2-70B & FP16 & W16A16 & 65.20 & 57.80 & 80.40 & 74.60 & 69.10 & 77.98 & 82.75 & 83.81 & 80.98 & 81.38 \\ 
 & SmoothQuant & W8A8 & 63.23 & 56.46 & 79.23 & 72.42 & 67.40 & 78.14 & 82.37 & 82.60 & 80.72 & 80.96 \\ 
 & GPTQ & W4A16 & 62.93 & 57.65 & 79.62 & 74.12 & 68.04 & 78.06 & 82.92 & 83.37 & 80.89 & 81.31 \\ 
 & FPTQ & W4A8 & 62.83 & 55.27 & 78.23 & 72.49 & 66.81 & 77.03 & \textbf{82.37} & 82.58 & 79.88 & 80.47 \\ 
\bottomrule
\end{tabular}
\caption{Comparison on MMLU and Common Sense QA. BW: BitWidth} 
\label{table:main table}
\end{table}

\subsection{Results on MMLU and Common Sense QA}

MMLU~\citep{hendrycks2020measuring} and Common Sense QA ~\citep{talmor2019commonsenseqa}  are currently renowned datasets that comprehensively reflect the performance of LLMs. We conducted extensive experiments on these datasets, including comparative assessments with two state-of-the-art solutions: SmoothQuant for W8A8, and GPTQ for W4A16.

On the MMLU dataset, our approach exhibits a performance gap within 1\% for most models compared to SmoothQuant. Notable outliers include LLaMA-7B and LLaMA-13B, which show a more pronounced drop. However, it's important to note that the MMLU dataset, with its predominant composition of multiple-choice questions, may exhibit bias in precision estimation when the inherent capabilities of the model are limited.

On Common Sense QA, our approach demonstrates a mere 1\% precision gap with the FP16 model across nearly all models, including the previously identified underperforming models LLaMA-7B and LLaMA-13B on MMLU. This observation underscores the robustness of our approach.


\begin{wraptable}{r}{8cm}
\vskip -0.1in
\setlength\tabcolsep{1pt}
\centering
\small
\begin{tabular}{@{}l*{7}{c}}
\toprule
Model & Method & Setting & WG & PIQA & HS & ARC$_e$ & Avg. \\ 
\midrule
LLaMA-7B & Original & FP16& 69.85 & 79.16 & 44.40 & 72.81 & 74.51\\
& LLM-QAT & W4A8 & 68.80 & 77.40 & 73.00 & 68.40 & 71.90\\
& FPTQ & W4A8 & \textbf{70.09} & \textbf{78.62} & \textbf{74.45} & \textbf{70.37} & \textbf{73.38}\\
\midrule
LLaMA-13B & Original & FP16 & 72.22 & 80.52 & 79.38 & 77.44 & 77.39 \\
& LLM-QAT & W4A8 & 70.60 & 79.10 & 77.50 & 73.00 & 75.05 \\
& FPTQ & W4A8 & \textbf{72.85} & \textbf{80.09} & \textbf{78.20} & \textbf{76.09} & \textbf{76.81}\\
\bottomrule
\end{tabular}
\caption{Comparison with LLM-QAT on LLaMA-7B.} 
\label{tab:LLaMA-7B-winogrande}
\end{wraptable}

\subsection{Comparison with LLM-QAT on Common Sense QA}

Given the paucity of other Post-training Quantization (PTQ) works employing W4A8 quantization, we conducted a comparative study with the Quantization-Aware Training (QAT) method, LLM-QAT~\citep{liu2023llmqat}, on the Common Sense QA dataset. Our approach achieved a precision that was notably closer to the FP16 model compared to LLM-QAT. However, due to the limited data publicly available from LLM-QAT, we present here the experimental results for only LLaMA-7B and LLaMA-13B. It can be observed that our approach yields slightly superior results on every subset of the dataset compared to LLM-QAT, highlighting the effectiveness of our methodology.

\section{Ablation Study}
\subsection{Comparison with Data-Free Quantization}
We acknowledge that the calibration dataset may be one of the factors affecting the performance of the quantized model. Therefore, to maintain fairness, we utilized the Pile dataset \citep{gao2020pile} as a calibration dataset in our previous experiments. However, to demonstrate the robustness of our method, we applied randomly generated tokens for model calibration. We conducted ablation studies on BLOOM-7B1, LLaMA-7B and LLaMA-2-7B under W8A8 and W4A8 bit-width settings in Table \ref{tab:Ablation study on Calib data}. It's exhilarating to note that, it was found that using a random dataset often resulted in superior results in most cases. This attests that our method is applicable in data-free situations.

\begin{table}[htbp]
\centering
\small
\setlength\tabcolsep{2pt}
\begin{tabular}{@{}l*{12}{c}}
\toprule
Model & \multicolumn{2}{c|}{HyperParam}  & \multicolumn{5}{c|}{MMLU} & \multicolumn{5}{c}{Common Sense QA} \\
& Calibration & BW & Hums. & STEM & Social & Other & Avg & WG & PIQA & HS & ARC$_e$ & Avg \\
\midrule
LLaMA-7B & Pile & W8A8 & 33.88 & 30.32 & 37.63 & 39.08 & 35.14 & 70.09 & 79.00 & 75.17 & 72.22 & 74.12 \\
 & Random & W8A8 & 32.33 & 29.85 & 36.46 & 38.25 & 34.07 & 70.01 & 78.62 & 75.48 & 72.69 & 74.20 \\ 
 & Pile & W4A8 & 30.20 & 29.95 & 32.76 & 35.87 & 32.02 & 70.01 & 78.40 & 74.46 & 70.79 & 73.42 \\
 & Random & W4A8 & 31.20 & 31.05 & 36.37 & 37.01 & 33.64 & 68.67 & 78.62 & 74.62 & 71.21 & 73.28 \\ 
\midrule
 LLaMA-2-7B & Pile & W8A8 & 42.49 & 36.65 & 50.67 & 51.33 & 45.06 & 69.06 & 77.97 & 75.91 & 75.98 & 74.58  \\
 & Random & W8A8 & 42.55 & 36.28 & 51.41 & 51.63 & 45.24 & 67.80 & 79.22 & 75.98 & 74.28 & 74.32 \\ 
 & Pile & W4A8 & 41.15 & 35.79 & 49.37 & 50.77 & 44.02 & 69.38 & 77.97 & 74.89 & 72.85 & 73.77  \\
 & Random & W4A8 & 41.32 & 35.42 & 47.97 & 50.19 & 43.56 & 67.88 & 78.07 & 75.46 & 73.11 & 73.63 \\ 
\midrule
BLOOM-7B1 & Pile & W8A8 & 26.04 & 27.80 & 24.50 & 25.82 & 26.03 & 63.93 & 72.91 &	57.24 & 57.74 &	62.96 \\ 
 & Random & W8A8 & 25.80 & 27.60 & 25.06 & 26.96 & 26.29 & 63.77 & 72.80 & 56.65 & 57.45 & 62.67 \\
 & Pile & W4A8 & 25.87 & 26.71 & 23.76 & 26.56 & 25.74 & 61.96 & 72.52 & 56.66 &	57.41 &	62.14 \\
 & Random & W4A8 & 26.29 & 27.04 & 23.37 & 27.08 & 25.99 & 61.88 & 72.42 & 56.17 & 56.94 & 61.85 \\
\bottomrule
\end{tabular}
\caption{Ablation study on calibration datasets on MMLU and Common Sense QA.} 
\label{tab:Ablation study on Calib data}
\end{table}

\subsection{Weight Quantization with GPTQ}

We observe that the GPTQ method, which compensates weights based on the Hessian matrix, is orthogonal to our existing approach. Therefore, we attempted to fine-tune the weights using the GPTQ method after conducting logarithmic activation equalization (LAE) on the model, to investigate the potential for increased precision. However, our experiments in Table~\ref{tab:Ablation on additional GPTQ} demonstrated that the addition of GPTQ operations generally resulted in a negative impact on precision in most cases. We encourage future researchers to conduct more intriguing explorations in this area.

\begin{table}[htbp]
\centering
\small
\setlength\tabcolsep{2pt}
\begin{tabular}{@{}l*{12}{c}}
\toprule
Model & \multicolumn{2}{c|}{HyperParam}  & \multicolumn{5}{c|}{MMLU} & \multicolumn{5}{c}{Common Sense QA} \\
& Method & BW & Hums. & STEM & Social & Other & Avg & WG & PIQA & HS & ARC$_e$ & Avg \\
\midrule
\multirow{3}{*}{LLaMA-7B} & FP16 & W16A16 & 33.60 & 31.10 & 38.20 & 38.40 & 35.20 & 69.85 & 79.16 & 76.21 & 72.81 & 74.51 \\
& FPTQ & W4A8 & 30.20 & 29.95 & 32.76 & 35.87 & 32.02 & 70.01 & 78.40 & 74.46 & 70.79 & 73.42 \\
& FPTQ$_{\text{GPTQ}}$ & W4A8  & 28.40 & 28.33 & 30.84 & 33.22 & 30.03 & 68.82 & 78.13 & 72.88 & 66.96 & 71.70 \\
\midrule
\multirow{3}{*}{LLaMA-2-7B} & FP16 & W16A16 & 43.40 & 37.00 & 51.80 & 52.40 & 46.00 & 69.06 & 79.11 & 75.98 & 74.58 & 74.68 \\ 
& FPTQ & W4A8 & 41.15 & 35.79 & 49.37 & 50.77 & 44.02 & 69.38 & 77.97 & 74.89 & 72.85 & 73.77 \\
& FPTQ$_{\text{GPTQ}}$ & W4A8  & 40.57 & 35.42 & 48.55 & 48.86 & 43.13 & 67.56 & 78.35 & 74.90 & 72.94 & 73.44 \\
\midrule
\multirow{3}{*}{BLOOM-7B1} & FP16 & W16A16 & 26.10 & 26.84 & 24.21 & 26.34 & 25.90 & 63.93 & 72.91 & 57.24 & 57.74 & 62.96 \\
 & FPTQ & W4A8 & 25.87 & 26.71 & 23.76 & 26.56 & 25.74 & 61.96 & 72.52 & 56.66 &	57.41 &	62.14 \\
& FPTQ$_{\text{GPTQ}}$ & W4A8  & 26.21 & 28.20 & 25.28 & 26.37 & 26.47 & 62.90 & 72.31 & 55.39 & 57.28 & 61.97 \\
\bottomrule
\end{tabular}
\caption{Ablation on MMLU and Common Sense QA. FPTQ$_{\text{GPTQ}}$: weights updated by GPTQ first.} 
\label{tab:Ablation on additional GPTQ}
\end{table}


\section{Discussion and Future Directions}
\textbf{Analysis on computation efficiency.} Modern GPUs, such as the NVIDIA A100, support parallel block-wise matrix computation and pipeline processing. Fine-grained weight quantization enjoys such block-wise computation and introduces little overhead. Currently, the W4A16  acceleration is based on the GPU \texttt{FP16INT4 GEMM} kernel~\citep{kim2022says}, which implements mixed-type matrix computation. The INT4 weights are first converted to FP16, and matrix computation is then performed with FP16. The underlying computation still uses the GPU's floating-point computation unit, so in the case of long inputs and large batches, the \texttt{FP16INT4} kernel even has a negative effect compared to direct FP16 computation because of the additional conversion. The W8A8 computation acceleration is based on the GPU \texttt{INT8 GEMM} kernel, which uses INT8 Tensor Cores for underlying computation. There is a noticeable acceleration in the context decoding stage, but in the self-decoding stage, the bottleneck mainly lies in memory access.

To simultaneously address the acceleration issues in both the context decoding and self-decoding stages, we can design an \texttt{INT8INT4} kernel, which profits INT8 Tensor Cores for acceleration in the context decoding stage, while keeping the weights loaded as INT4 to reduce memory access time in the self-decoding stage.



\textbf{Data-free quantization.} We discover that it is promising to randomly draw samples from the token vocabulary as in Table~\ref{tab:Ablation study on Calib data}. We believe that there is still room for improvement in this regard.

\textbf{Scale computation requires activation only.} For activation quantization, our method completely removes weights for the computation of $\mathbf{s}_i$ which echoes the findings in \cite{lin2023awq}. To make our strategy more generalizable, we introduce a hyper-parameter $\alpha$ to control the level of suppression, see \ref{app:lae-general}. It is however possible to devise other non-linear mapping functions that are hyper-parameter free and in the meanwhile lead to better performance.

\section{Conclusion}
In conclusion, our work presents a significant stride in the domain of Large Language Model (LLM) compression. Upon an overview of the existing quantization schemes, we introduce a novel post-training quantization approach that can make the inference of LLMs more efficient, without compromising their performance.
We successfully achieved high performance and efficiency for W4A8, which has the optimal utilization of computational resources which enhances the speed of both content-decoding and self-decoding stages.
Furthermore, the removal of the need for fine-tuning during the training process simplifies the deployment pipeline significantly. This attests that our method provides an effective deployable solution for LLMs without sacrificing their accuracy. While our progress is encouraging, we acknowledge the potential for further exploration and refinement in this area. We anticipate that our work will inspire future research endeavors aimed at making LLMs even more efficient and practical.


\newpage
\bibliography{iclr2023_conference}

\begin{thebibliography}{41}
\providecommand{\natexlab}[1]{#1}
\providecommand{\url}[1]{\texttt{#1}}
\expandafter\ifx\csname urlstyle\endcsname\relax
  \providecommand{\doi}[1]{doi: #1}\else
  \providecommand{\doi}{doi: \begingroup \urlstyle{rm}\Url}\fi

\bibitem[Ba et~al.(2016)Ba, Kiros, and Hinton]{ba2016layer}
Jimmy~Lei Ba, Jamie~Ryan Kiros, and Geoffrey~E Hinton.
\newblock Layer normalization.
\newblock \emph{arXiv preprint arXiv:1607.06450}, 2016.

\bibitem[Brown et~al.(2020)Brown, Mann, Ryder, Subbiah, Kaplan, Dhariwal,
  Neelakantan, Shyam, Sastry, Askell, et~al.]{brown2020language}
Tom Brown, Benjamin Mann, Nick Ryder, Melanie Subbiah, Jared~D Kaplan, Prafulla
  Dhariwal, Arvind Neelakantan, Pranav Shyam, Girish Sastry, Amanda Askell,
  et~al.
\newblock Language models are few-shot learners.
\newblock In \emph{Conference on Neural Information Processing Systems
  (NeurIPS)}, 2020.

\bibitem[Dettmers et~al.(2022)Dettmers, Lewis, Belkada, and
  Zettlemoyer]{dettmers2022llm}
Tim Dettmers, Mike Lewis, Younes Belkada, and Luke Zettlemoyer.
\newblock {LLM.int8()}: 8-bit matrix multiplication for transformers at scale.
\newblock \emph{arXiv preprint arXiv:2208.07339}, 2022.

\bibitem[Devlin et~al.(2019)Devlin, Chang, Lee, and Toutanova]{devlin2018bert}
Jacob Devlin, Ming-Wei Chang, Kenton Lee, and Kristina Toutanova.
\newblock {BERT}: Pre-training of deep bidirectional transformers for language
  understanding.
\newblock In \emph{North American Chapter of the Association for Computational
  Linguistics (NAACL)}, 2019.

\bibitem[Du et~al.(2021)Du, Qian, Liu, Ding, Qiu, Yang, and Tang]{du2021glm}
Zhengxiao Du, Yujie Qian, Xiao Liu, Ming Ding, Jiezhong Qiu, Zhilin Yang, and
  Jie Tang.
\newblock Glm: General language model pretraining with autoregressive blank
  infilling.
\newblock \emph{arXiv preprint arXiv:2103.10360}, 2021.

\bibitem[Frantar \& Alistarh(2023)Frantar and Alistarh]{frantar2023sparsegpt}
Elias Frantar and Dan Alistarh.
\newblock Sparsegpt: Massive language models can be accurately pruned in
  one-shot, 2023.

\bibitem[Frantar et~al.(2022)Frantar, Ashkboos, Hoefler, and
  Alistarh]{frantar2022gptq}
Elias Frantar, Saleh Ashkboos, Torsten Hoefler, and Dan Alistarh.
\newblock Gptq: Accurate post-training quantization for generative pre-trained
  transformers.
\newblock \emph{arXiv preprint arXiv:2210.17323}, 2022.

\bibitem[Gao et~al.(2020)Gao, Biderman, Black, Golding, Hoppe, Foster, Phang,
  He, Thite, Nabeshima, et~al.]{gao2020pile}
Leo Gao, Stella Biderman, Sid Black, Laurence Golding, Travis Hoppe, Charles
  Foster, Jason Phang, Horace He, Anish Thite, Noa Nabeshima, et~al.
\newblock The pile: An 800gb dataset of diverse text for language modeling.
\newblock \emph{arXiv preprint arXiv:2101.00027}, 2020.

\bibitem[Gao et~al.(2021)Gao, Tow, Biderman, Black, DiPofi, Foster, Golding,
  Hsu, McDonell, Muennighoff, Phang, Reynolds, Tang, Thite, Wang, Wang, and
  Zou]{eval-harness}
Leo Gao, Jonathan Tow, Stella Biderman, Sid Black, Anthony DiPofi, Charles
  Foster, Laurence Golding, Jeffrey Hsu, Kyle McDonell, Niklas Muennighoff,
  Jason Phang, Laria Reynolds, Eric Tang, Anish Thite, Ben Wang, Kevin Wang,
  and Andy Zou.
\newblock A framework for few-shot language model evaluation, September 2021.
\newblock URL \url{https://doi.org/10.5281/zenodo.5371628}.

\bibitem[Hendrycks et~al.(2020)Hendrycks, Burns, Basart, Zou, Mazeika, Song,
  and Steinhardt]{hendrycks2020measuring}
Dan Hendrycks, Collin Burns, Steven Basart, Andy Zou, Mantas Mazeika, Dawn
  Song, and Jacob Steinhardt.
\newblock Measuring massive multitask language understanding.
\newblock \emph{arXiv preprint arXiv:2009.03300}, 2020.

\bibitem[Kim et~al.(2022)Kim, Henry, Fahim, and Awadalla]{kim2022says}
Young~Jin Kim, Rawn Henry, Raffy Fahim, and Hany~Hassan Awadalla.
\newblock Who says elephants can't run: Bringing large scale moe models into
  cloud scale production.
\newblock \emph{arXiv preprint arXiv:2211.10017}, 2022.

\bibitem[Lauren{\c{c}}on et~al.(2022)Lauren{\c{c}}on, Saulnier, Wang, Akiki,
  del Moral, Le~Scao, Von~Werra, Mou, Ponferrada, Nguyen,
  et~al.]{laurencconbigscience}
Hugo Lauren{\c{c}}on, Lucile Saulnier, Thomas Wang, Christopher Akiki,
  Albert~Villanova del Moral, Teven Le~Scao, Leandro Von~Werra, Chenghao Mou,
  Eduardo~Gonz{\'a}lez Ponferrada, Huu Nguyen, et~al.
\newblock The {BigScience} corpus: A 1.6 {TB} composite multilingual dataset.
\newblock 2022.

\bibitem[Lin et~al.(2023)Lin, Tang, Tang, Yang, Dang, and Han]{lin2023awq}
Ji~Lin, Jiaming Tang, Haotian Tang, Shang Yang, Xingyu Dang, and Song Han.
\newblock Awq: Activation-aware weight quantization for llm compression and
  acceleration, 2023.

\bibitem[Liu et~al.(2019)Liu, Ott, Goyal, Du, Joshi, Chen, Levy, Lewis,
  Zettlemoyer, and Stoyanov]{liu2019roberta}
Yinhan Liu, Myle Ott, Naman Goyal, Jingfei Du, Mandar Joshi, Danqi Chen, Omer
  Levy, Mike Lewis, Luke Zettlemoyer, and Veselin Stoyanov.
\newblock Roberta: A robustly optimized bert pretraining approach.
\newblock \emph{arXiv preprint arXiv:1907.11692}, 2019.

\bibitem[Liu et~al.(2023{\natexlab{a}})Liu, Oguz, Zhao, Chang, Stock, Mehdad,
  Shi, Krishnamoorthi, and Chandra]{liu2023llm}
Zechun Liu, Barlas Oguz, Changsheng Zhao, Ernie Chang, Pierre Stock, Yashar
  Mehdad, Yangyang Shi, Raghuraman Krishnamoorthi, and Vikas Chandra.
\newblock Llm-qat: Data-free quantization aware training for large language
  models.
\newblock \emph{arXiv preprint arXiv:2305.17888}, 2023{\natexlab{a}}.

\bibitem[Liu et~al.(2023{\natexlab{b}})Liu, Oguz, Zhao, Chang, Stock, Mehdad,
  Shi, Krishnamoorthi, and Chandra]{liu2023llmqat}
Zechun Liu, Barlas Oguz, Changsheng Zhao, Ernie Chang, Pierre Stock, Yashar
  Mehdad, Yangyang Shi, Raghuraman Krishnamoorthi, and Vikas Chandra.
\newblock Llm-qat: Data-free quantization aware training for large language
  models, 2023{\natexlab{b}}.

\bibitem[Ma et~al.(2023)Ma, Fang, and Wang]{ma2023llmpruner}
Xinyin Ma, Gongfan Fang, and Xinchao Wang.
\newblock Llm-pruner: On the structural pruning of large language models, 2023.

\bibitem[Nagel et~al.(2019)Nagel, Baalen, Blankevoort, and
  Welling]{nagel2019data}
Markus Nagel, Mart~van Baalen, Tijmen Blankevoort, and Max Welling.
\newblock Data-free quantization through weight equalization and bias
  correction.
\newblock In \emph{International Conference on Computer Vision (ICCV)}, 2019.

\bibitem[Nagel et~al.(2021)Nagel, Fournarakis, Amjad, Bondarenko, van Baalen,
  and Blankevoort]{nagel2021white}
Markus Nagel, Marios Fournarakis, Rana~Ali Amjad, Yelysei Bondarenko, Mart van
  Baalen, and Tijmen Blankevoort.
\newblock A white paper on neural network quantization.
\newblock \emph{arXiv preprint arXiv:2106.08295}, 2021.

\bibitem[Paperno et~al.(2016)Paperno, Kruszewski, Lazaridou, Pham, Bernardi,
  Pezzelle, Baroni, Boleda, and Fern{\'a}ndez]{paperno2016lambada}
Denis Paperno, Germ{\'a}n Kruszewski, Angeliki Lazaridou, Quan~Ngoc Pham,
  Raffaella Bernardi, Sandro Pezzelle, Marco Baroni, Gemma Boleda, and Raquel
  Fern{\'a}ndez.
\newblock The {LAMBADA} dataset: Word prediction requiring a broad discourse
  context.
\newblock \emph{arXiv preprint arXiv:1606.06031}, 2016.

\bibitem[Raffel et~al.(2020)Raffel, Shazeer, Roberts, Lee, Narang, Matena,
  Zhou, Li, and Liu]{raffel2020exploring}
Colin Raffel, Noam Shazeer, Adam Roberts, Katherine Lee, Sharan Narang, Michael
  Matena, Yanqi Zhou, Wei Li, and Peter~J Liu.
\newblock Exploring the limits of transfer learning with a unified text-to-text
  transformer.
\newblock \emph{The Journal of Machine Learning Research}, 21\penalty0
  (1):\penalty0 5485--5551, 2020.

\bibitem[Sakaguchi et~al.(2021)Sakaguchi, Bras, Bhagavatula, and
  Choi]{sakaguchi2021winogrande}
Keisuke Sakaguchi, Ronan~Le Bras, Chandra Bhagavatula, and Yejin Choi.
\newblock Winogrande: An adversarial winograd schema challenge at scale.
\newblock \emph{Communications of the ACM}, 64\penalty0 (9):\penalty0 99--106,
  2021.

\bibitem[Scao et~al.(2022)Scao, Fan, Akiki, Pavlick, Ili{\'c}, Hesslow,
  Castagn{\'e}, Luccioni, Yvon, Gall{\'e}, et~al.]{scao2022bloom}
Teven~Le Scao, Angela Fan, Christopher Akiki, Ellie Pavlick, Suzana Ili{\'c},
  Daniel Hesslow, Roman Castagn{\'e}, Alexandra~Sasha Luccioni, Fran{\c{c}}ois
  Yvon, Matthias Gall{\'e}, et~al.
\newblock Bloom: A 176b-parameter open-access multilingual language model.
\newblock \emph{arXiv preprint arXiv:2211.05100}, 2022.

\bibitem[Shen et~al.(2020)Shen, Dong, Ye, Ma, Yao, Gholami, Mahoney, and
  Keutzer]{shen2020q}
Sheng Shen, Zhen Dong, Jiayu Ye, Linjian Ma, Zhewei Yao, Amir Gholami,
  Michael~W Mahoney, and Kurt Keutzer.
\newblock Q-bert: Hessian based ultra low precision quantization of bert.
\newblock In \emph{Proceedings of the AAAI Conference on Artificial
  Intelligence}, volume~34, pp.\  8815--8821, 2020.

\bibitem[Sun et~al.(2023)Sun, Liu, Bair, and Kolter]{sun2023simple}
Mingjie Sun, Zhuang Liu, Anna Bair, and J~Zico Kolter.
\newblock A simple and effective pruning approach for large language models.
\newblock \emph{arXiv preprint arXiv:2306.11695}, 2023.

\bibitem[Talmor et~al.(2019)Talmor, Herzig, Lourie, and
  Berant]{talmor2019commonsenseqa}
Alon Talmor, Jonathan Herzig, Nicholas Lourie, and Jonathan Berant.
\newblock Commonsenseqa: A question answering challenge targeting commonsense
  knowledge, 2019.

\bibitem[Tata \& Patel(2003)Tata and Patel]{tata2003piqa}
Sandeep Tata and Jignesh~M Patel.
\newblock {PiQA}: An algebra for querying protein data sets.
\newblock In \emph{International Conference on Scientific and Statistical
  Database Management}, 2003.

\bibitem[Touvron et~al.(2023)Touvron, Lavril, Izacard, Martinet, Lachaux,
  Lacroix, Rozi{\`e}re, Goyal, Hambro, Azhar, et~al.]{touvron2023llama}
Hugo Touvron, Thibaut Lavril, Gautier Izacard, Xavier Martinet, Marie-Anne
  Lachaux, Timoth{\'e}e Lacroix, Baptiste Rozi{\`e}re, Naman Goyal, Eric
  Hambro, Faisal Azhar, et~al.
\newblock Llama: Open and efficient foundation language models.
\newblock \emph{arXiv preprint arXiv:2302.13971}, 2023.

\bibitem[Vaswani et~al.(2017)Vaswani, Shazeer, Parmar, Uszkoreit, Jones, Gomez,
  Kaiser, and Polosukhin]{vaswani2017attention}
Ashish Vaswani, Noam Shazeer, Niki Parmar, Jakob Uszkoreit, Llion Jones,
  Aidan~N Gomez, {\L}ukasz Kaiser, and Illia Polosukhin.
\newblock Attention is all you need.
\newblock In \emph{Conference on Neural Information Processing Systems
  (NeurIPS)}, 2017.

\bibitem[Wei et~al.(2022)Wei, Tay, Bommasani, Raffel, Zoph, Borgeaud, Yogatama,
  Bosma, Zhou, Metzler, et~al.]{wei2022emergent}
Jason Wei, Yi~Tay, Rishi Bommasani, Colin Raffel, Barret Zoph, Sebastian
  Borgeaud, Dani Yogatama, Maarten Bosma, Denny Zhou, Donald Metzler, et~al.
\newblock Emergent abilities of large language models.
\newblock \emph{arXiv preprint arXiv:2206.07682}, 2022.

\bibitem[Wu et~al.(2020)Wu, Judd, Zhang, Isaev, and
  Micikevicius]{wu2020integer}
Hao Wu, Patrick Judd, Xiaojie Zhang, Mikhail Isaev, and Paulius Micikevicius.
\newblock Integer quantization for deep learning inference: Principles and
  empirical evaluation, 2020.

\bibitem[Wu et~al.(2023)Wu, Yao, and He]{wu2023zeroquantfp}
Xiaoxia Wu, Zhewei Yao, and Yuxiong He.
\newblock Zeroquant-fp: A leap forward in llms post-training w4a8 quantization
  using floating-point formats, 2023.

\bibitem[Xiao et~al.(2023)Xiao, Lin, Seznec, Wu, Demouth, and
  Han]{xiao2023smoothquant}
Guangxuan Xiao, Ji~Lin, Mickael Seznec, Hao Wu, Julien Demouth, and Song Han.
\newblock Smoothquant: Accurate and efficient post-training quantization for
  large language models.
\newblock In \emph{International Conference on Machine Learning}, pp.\
  38087--38099. PMLR, 2023.

\bibitem[Yang et~al.(2019)Yang, Dai, Yang, Carbonell, Salakhutdinov, and
  Le]{yang2019xlnet}
Zhilin Yang, Zihang Dai, Yiming Yang, Jaime Carbonell, Russ~R Salakhutdinov,
  and Quoc~V Le.
\newblock Xlnet: Generalized autoregressive pretraining for language
  understanding.
\newblock \emph{Advances in neural information processing systems}, 32, 2019.

\bibitem[Yao et~al.(2021)Yao, Dong, Zheng, Gholami, Yu, Tan, Wang, Huang, Wang,
  Mahoney, et~al.]{yao2021hawq}
Zhewei Yao, Zhen Dong, Zhangcheng Zheng, Amir Gholami, Jiali Yu, Eric Tan,
  Leyuan Wang, Qijing Huang, Yida Wang, Michael Mahoney, et~al.
\newblock {HAWQ-v3}: Dyadic neural network quantization.
\newblock In \emph{International Conference on Machine Learning (ICML)}, 2021.

\bibitem[Yao et~al.(2022)Yao, Aminabadi, Zhang, Wu, Li, and
  He]{yao2022zeroquant}
Zhewei Yao, Reza~Yazdani Aminabadi, Minjia Zhang, Xiaoxia Wu, Conglong Li, and
  Yuxiong He.
\newblock {ZeroQuant}: Efficient and affordable post-training quantization for
  large-scale transformers.
\newblock \emph{arXiv preprint arXiv:2206.01861}, 2022.

\bibitem[Yao et~al.(2023)Yao, Wu, Li, Youn, and He]{yao2023zeroquantv2}
Zhewei Yao, Xiaoxia Wu, Cheng Li, Stephen Youn, and Yuxiong He.
\newblock Zeroquant-v2: Exploring post-training quantization in llms from
  comprehensive study to low rank compensation, 2023.

\bibitem[Yuan et~al.(2023)Yuan, Niu, Liu, Liu, Wang, Shang, Sun, Wu, Wu, and
  Wu]{yuan2023rptq}
Zhihang Yuan, Lin Niu, Jiawei Liu, Wenyu Liu, Xinggang Wang, Yuzhang Shang,
  Guangyu Sun, Qiang Wu, Jiaxiang Wu, and Bingzhe Wu.
\newblock Rptq: Reorder-based post-training quantization for large language
  models, 2023.

\bibitem[Zellers et~al.(2019)Zellers, Holtzman, Bisk, Farhadi, and
  Choi]{zellers2019hellaswag}
Rowan Zellers, Ari Holtzman, Yonatan Bisk, Ali Farhadi, and Yejin Choi.
\newblock Hellaswag: Can a machine really finish your sentence?
\newblock \emph{arXiv preprint arXiv:1905.07830}, 2019.

\bibitem[Zhang et~al.(2023)Zhang, Yang, Liu, Wang, Xian, Wang, and
  Song]{zhang2023lifting}
Chen Zhang, Yang Yang, Jiahao Liu, Jingang Wang, Yunsen Xian, Benyou Wang, and
  Dawei Song.
\newblock Lifting the curse of capacity gap in distilling language models,
  2023.

\bibitem[Zhang et~al.(2022)Zhang, Roller, Goyal, Artetxe, Chen, Chen, Dewan,
  Diab, Li, Lin, et~al.]{zhang2022opt}
Susan Zhang, Stephen Roller, Naman Goyal, Mikel Artetxe, Moya Chen, Shuohui
  Chen, Christopher Dewan, Mona Diab, Xian Li, Xi~Victoria Lin, et~al.
\newblock {OPT}: Open pre-trained transformer language models.
\newblock \emph{arXiv preprint arXiv:2205.01068}, 2022.

\end{thebibliography}
\bibliographystyle{iclr2023_conference}

\newpage

\appendix
\section{Appendix}

\subsection{Preliminary Knowledge on Quantization}

Quantization is a process of mapping continuous values to discrete ones by scaling. The scaling factor is also called quantization step size. In practice, a higher-precision floating point is used for training and the quantized version is used for inference. Consider $b$-bit integer quantization, for a real tensor $\mathbf{x}$ ranges in $(min, max)$,  it can be converted to an integer tensor $\mathbf{x}'$ within $(-2^{b-1}, 2^{b-1}-1)$ by symmetric uniform quantization as,

\begin{align}\label{eq:quant-scale}
    scale &= max(\mathbf{|x|})/(2^{b-1} -1 ) \\
    \mathbf{x}' &= \round{(\mathbf{x} / scale)}
\end{align}

\textbf{Weight quantization and activation quantization.} Typically, the weight is quantized as integer values. Activation quantization refers to the quantization of intermediate activation feature maps.

\textbf{Static quantization vs. dynamic quantization.} For static quantization, offline activation statistics are collected to compute the scale and it is kept static during inference. For dynamic quantization, such statistics are computed at runtime.

\textbf{Per-tensor vs. per-token.} In the per-tensor scheme, the tensor matrix is considered as a whole to compute the quantization scale. In the per-token scheme, each input token corresponds to a scale computed upon all activation channels of the specific token. In essence, the per-token scheme is more fine-grained.

\textbf{Per-channel vs. group-wise.} In the per-channel scheme, the quantization scale is computed channel-wise. In the group-wise scheme, each channel is divided into several groups and so are its scales.

\subsection{Generalized Form of LAE}\label{app:lae-general}

We give a generalized form of logarithmic activation equalization function. For the majority of LLMs, we use $\alpha=1$. 

\begin{equation}\label{eq:log-act-eq_appendix}
    scale = (\log_2 (2+scale))^{\alpha}
\end{equation}

\subsection{More Activation Distribution of LLMs}~\label{app:more-act-dist}

\begin{figure*}[htbp]
  \centering
    \includegraphics[width=\textwidth, trim=200 0 200 0]{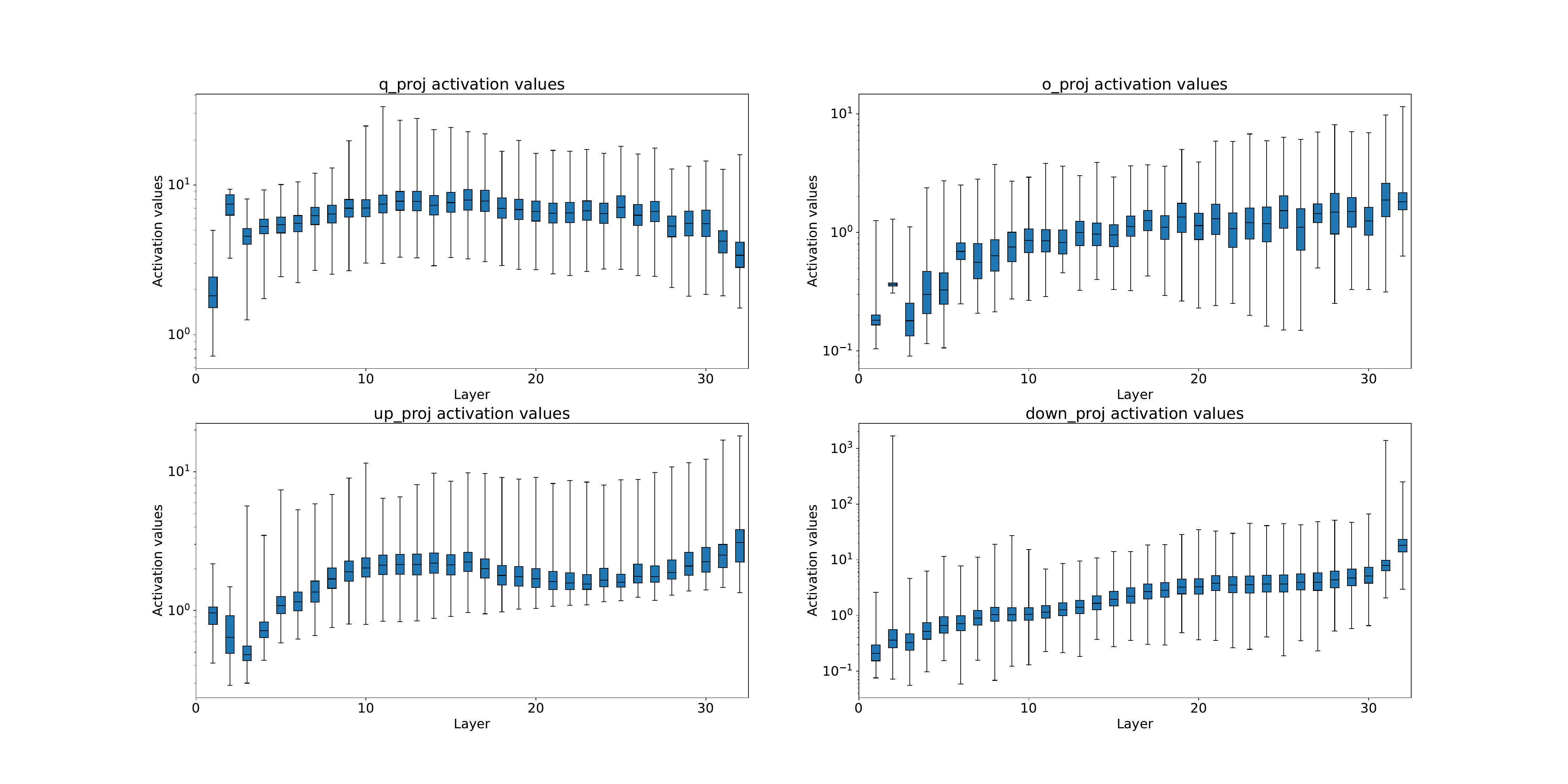}
  \caption{Visualization of activation distribution of $o_{proj}$ and $down_{proj}$ on LLaMA-2-7B.}
  \label{fig:activation distribution on LLMs (LLaMA-2-7B)}
\end{figure*}

\begin{figure*}[htbp]
  \centering
    \includegraphics[width=\textwidth, trim=200 0 200 0]{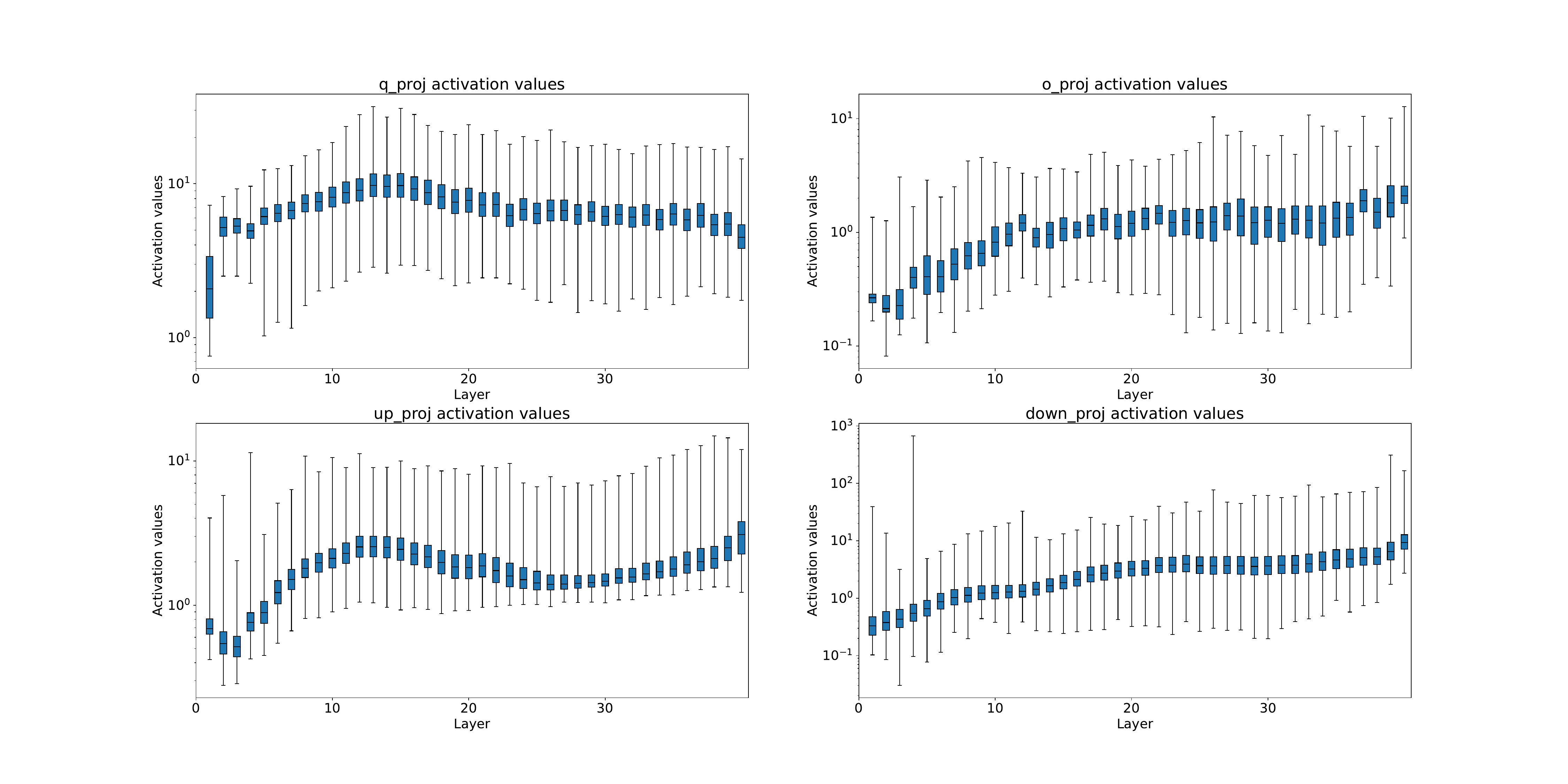}
  \caption{Visualization of activation distribution of $o_{proj}$ and $down_{proj}$ on LLaMA-2-13B.}
  \label{fig:activation distribution on LLMs (LLaMA-2-13B)}
\end{figure*}

\begin{figure*}[htbp]
  \centering
    \includegraphics[width=\textwidth, trim=200 0 200 0]{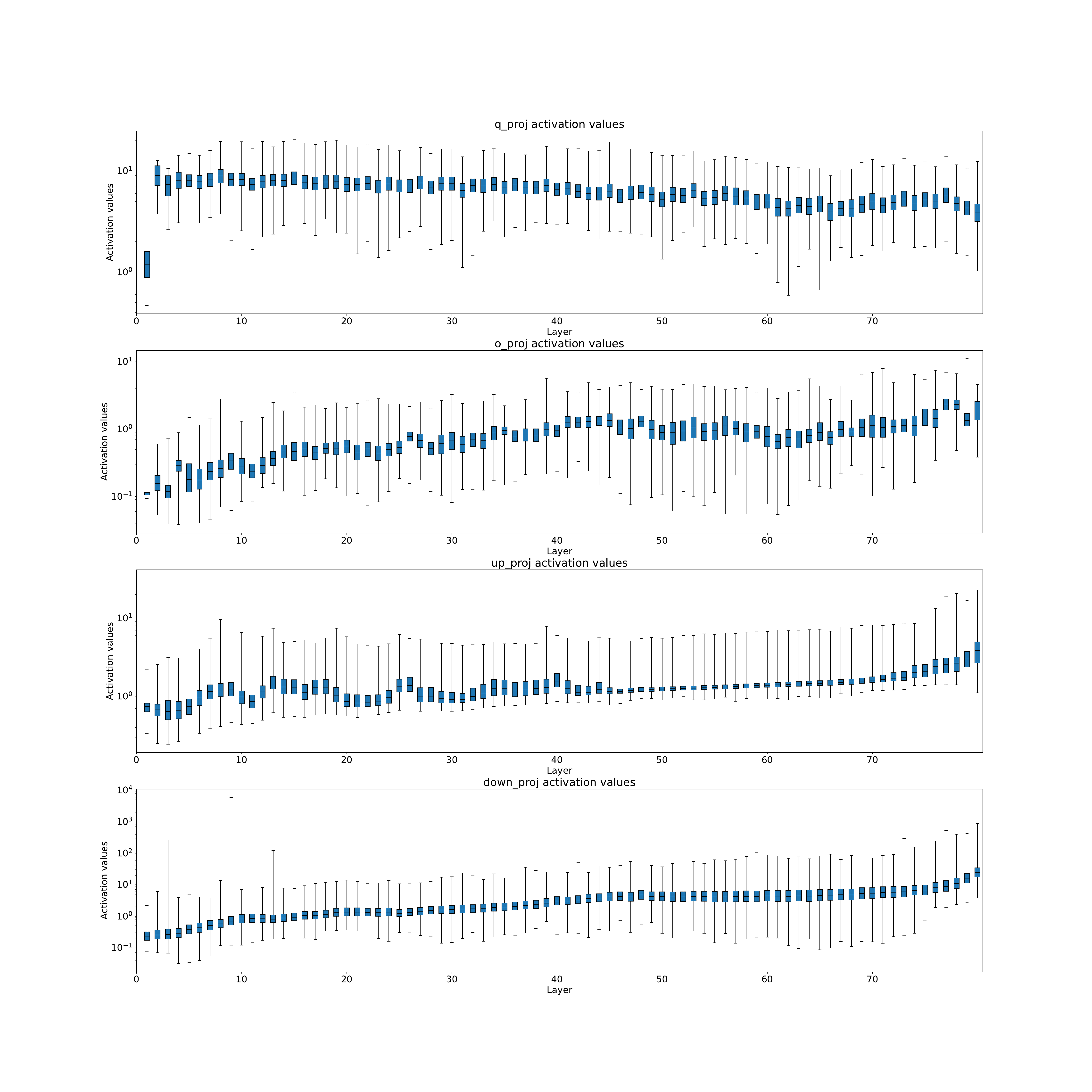}
  \caption{Visualization of activation distribution of $o_{proj}$ and $down_{proj}$ on LLaMA-2-70B.}
  \label{fig:activation distribution on LLMs (LLaMA-2-70B)}
\end{figure*}

\begin{figure*}[htbp]
  \centering
    \includegraphics[width=\textwidth, trim=200 0 200 0]{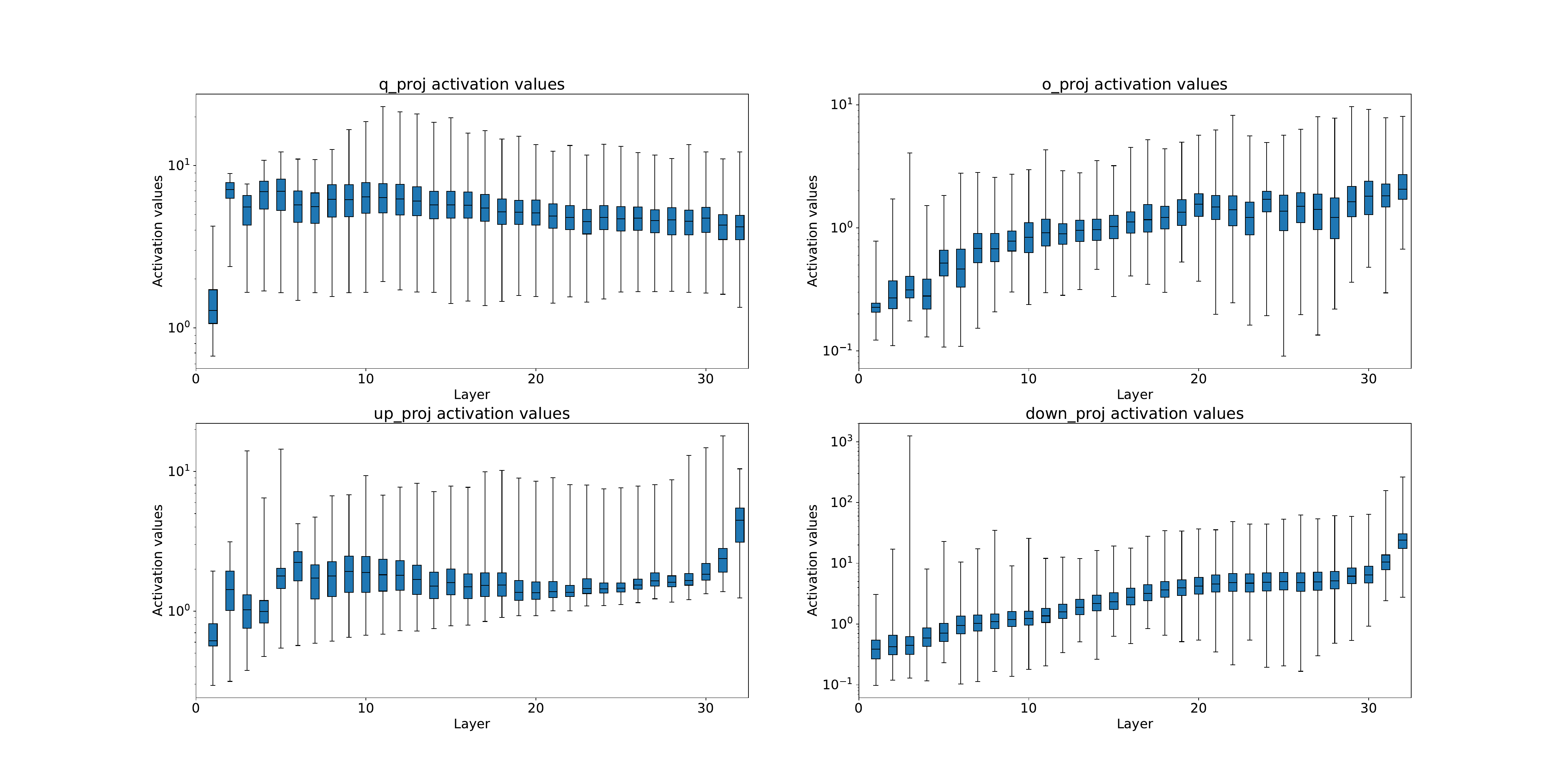}
  \caption{Visualization of activation distribution of $o_{proj}$ and $down_{proj}$ on LLaMA-7B.}
  \label{fig:activation distribution on LLMs (LLaMA-7B)}
\end{figure*}

\begin{figure*}[htbp]
  \centering
    \includegraphics[width=\textwidth, trim=200 0 200 0]{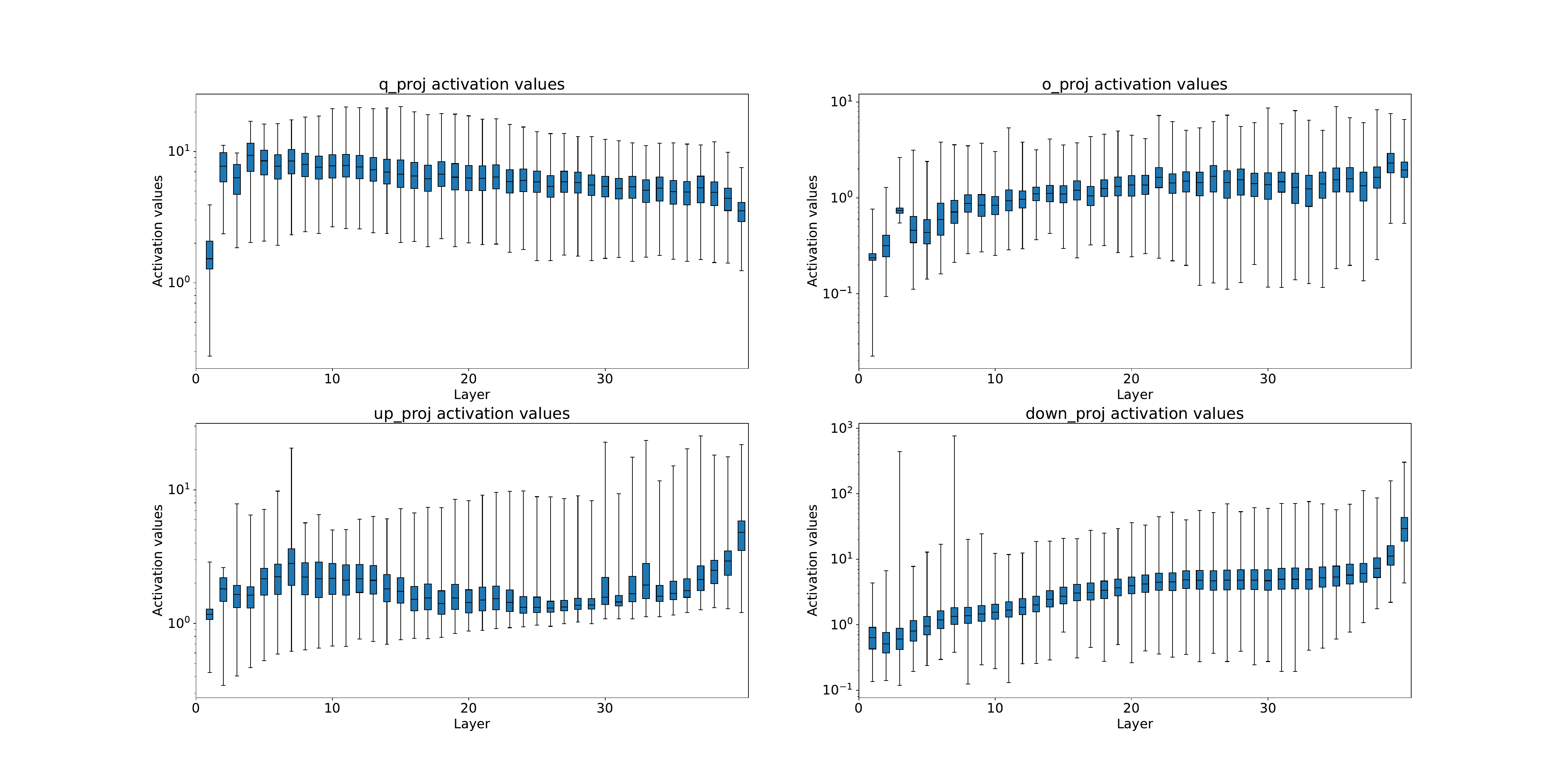}
  \caption{Visualization of activation distribution of $o_{proj}$ and $down_{proj}$ on LLaMA-13B.}
  \label{fig:activation distribution on LLMs (LLaMA-13B)}
\end{figure*}

\begin{figure*}[htbp]
  \centering
    \includegraphics[width=\textwidth, trim=200 0 200 0]{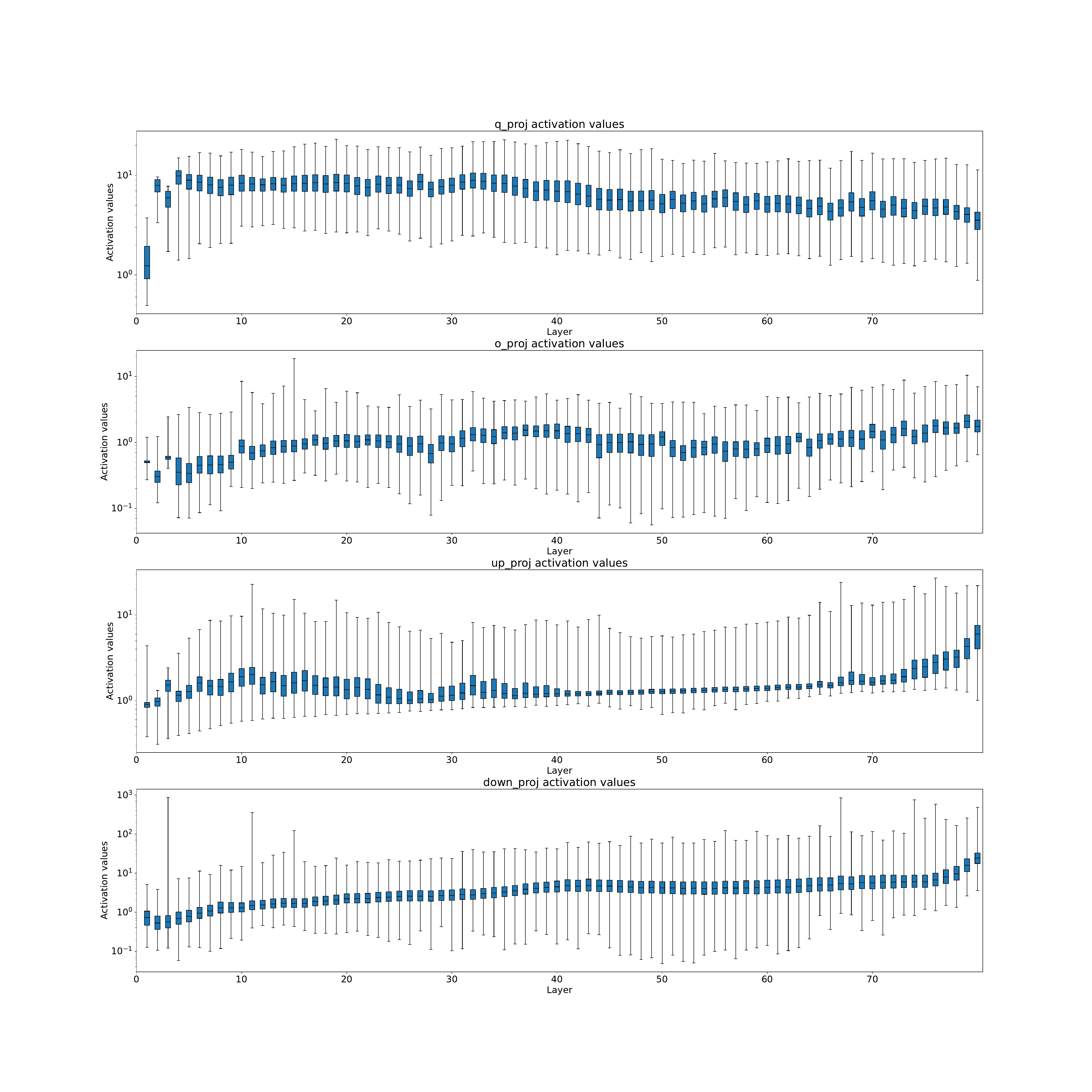}
  \caption{Visualization of activation distribution of $o_{proj}$ and $down_{proj}$ on LLaMA-65B.}
  \label{fig:activation distribution on LLMs (LLaMA-65B)}
\end{figure*}

We visualize the activation distributions of the LLaMA series in Figure~\ref{fig:activation distribution on LLMs (LLaMA-2-7B)}, \ref{fig:activation distribution on LLMs (LLaMA-2-13B)}, \ref{fig:activation distribution on LLMs (LLaMA-2-70B)}, \ref{fig:activation distribution on LLMs (LLaMA-7B)}, \ref{fig:activation distribution on LLMs (LLaMA-13B)}, and \ref{fig:activation distribution on LLMs (LLaMA-65B)}. It is rather exciting to find that LLaMA models at different scales share similar distributions in the same operations, which leads to a universal quantization scheme.

\end{document}